\documentclass[runningheads]{llncs}
\usepackage[T1]{fontenc}
\usepackage{graphicx}
\usepackage{capt-of}
\usepackage{tabularx}
\usepackage{booktabs}
\usepackage[usenames,dvipsnames]{xcolor}
\usepackage{array}
\usepackage{hyperref}
\usepackage{pgfplots}
\pgfplotsset{compat=1.17}
\usepackage{color}

\hypersetup{
    colorlinks=true,
    linkcolor=blue,
    citecolor=blue,
    filecolor=blue,
    urlcolor=blue
}
\usepackage{amsmath}
\usepackage{amssymb}
\usepackage{comment}
\usepackage{tikz}
\usepackage{float}
\usepackage{tablefootnote}
\usepackage{multirow}
\usepackage[misc]{ifsym}

\begin{document}
\title{LORTSAR: Low-Rank Transformer for Skeleton-based Action Recognition}
\titlerunning{LORTSAR: Low-Rank Transformer for Skeleton-based Action Recognition}
%
\author{Soroush Oraki\hspace{0.2mm}\textsuperscript{(\hspace{0.15mm}\Letter\hspace{0.2mm})} \and
Harry Zhuang \and
Jie Liang}
\authorrunning{S. Oraki et al.}
%
\institute{School of Engineering Science, Simon Fraser University, Burnaby, BC, Canada\\
\email{\{soroush\_oraki,harry\_zhuang\_2,jiel\}@sfu.ca}}
\maketitle              
\begin{abstract}
The complexity of state-of-the-art Transformer-based models for skeleton-based action recognition poses significant challenges in terms of computational efficiency and resource utilization. In this paper, we explore the application of Singular Value Decomposition (SVD) to effectively reduce the model sizes of these pre-trained models, aiming to minimize their resource consumption while preserving accuracy. Our method, \texttt{LORTSAR} (LOw-Rank Transformer for Skeleton-based Action Recognition), also includes a fine-tuning step to compensate for any potential accuracy degradation caused by model compression, and is applied to two leading Transformer-based models, "Hyperformer" and "STEP-CATFormer". Experimental results on the "NTU RGB+D" and "NTU RGB+D 120" datasets show that our method can reduce the number of model parameters substantially with negligible degradation or even performance increase in recognition accuracy. This confirms that SVD combined with post-compression fine-tuning can boost model efficiency, paving the way for more sustainable, lightweight, and high-performance technologies in human action recognition.

\end{abstract}

\section{Introduction}

Skeleton-based human action recognition~\cite{ren2024survey,wang2023comprehensive,xin2023transformer} is a rapidly evolving field within computer vision, with many applications such as security, robotics, and healthcare \cite{SkeletonFall}. While early methods~\cite{du2015hierarchical,song2016endtoend,Ke_2017,duan2022revisiting} utilized RGB videos, their enormous storage requirements and sensitivity to environmental factors led to the adoption of skeleton-based methods, representing the human body as a set of interconnected joints and bones. Additionally, skeleton data can be easily matched to a graph input with the joints and bones depicted as the edges and vertices. This shift facilitated the use of Graph Convolutional Networks (GCNs)~\cite{stgcn,shi2019twostream,chen2021channelwise,Chi_2022_CVPR}, which excel at capturing the inherent structure of human pose. However, GCNs face challenges in modeling relationships between distant joints, have fixed topology, and are sensitive to noise, prompting ongoing research into alternative approaches.

Transformer models, leveraging attention mechanisms, has emerged as a powerful alternative, capable of capturing long-range dependencies and contextual information in skeleton sequences~\cite{vaswani2017,plizzari2021skeleton,zhou2023hypergraph,long2023step}. Although Transformer models can resolve several key issues of GCNs, they have their own limitations such as requiring adaptation of vanilla Transformers to skeleton data, necessitating positional embedding, and designing specialized attention modules. Consequently, most state-of-the-art (SOTA) Transformers are computationally intensive, often surpassing the very GCN models they aim to outperform~\cite{brown2020language,touvron2023llama}.

In response to these resource constraints, we propose applying the principles of low-rank approximation, specifically a selective rank reduction method using Singular Value Decomposition (SVD), to the largest weight matrices in SOTA Transformer models~\cite{frankle2019lottery,molchanov2017pruning,LRD,sharma2023truth}. Our experiments reveal that these matrices can be substantially rank reduced while maintaining high performance. Furthermore, fine-tuning Transformers with these reduced matrices can further enhance their performance, achieving a balance between accuracy and computational efficiency.

Our main contributions can be summarized as follows:
\begin{itemize}
  \item We introduce low-rank approximation to the SOTA Transformer models for skeleton-based action recognition to significantly reduce the number of parameters and decrease their computational demands.
  \item We test the performance of the proposed \texttt{LORTSAR} models with various levels of rank reduction to determine the most effective configurations.
  \item We retrain the Transformer-based models after applying low rank approximation with different rank reductions to determine potential further enhancements on performance.
\end{itemize}

\section{Related Work}

In this section, we discuss some related works which represent the stages of development for skeleton-based human action recognition as well as  model compression and low-rank approximation.

\subsection{Skeleton-based Action Recognition}

\subsubsection{RNN/CNN-based methods.}

Recurrent Neural Networks (RNNs)~\cite{du2015hierarchical,song2016endtoend} and Convolutional Neural Networks (CNNs)~\cite{Ke_2017,duan2022revisiting} are foundational in human action recognition research. RNNs effectively process sequential video data for temporal feature extraction, while CNNs excel at capturing spatial features from individual frames. However, their limitations in modeling spatial relationships between joints and capturing global context prompted the exploration of alternative deep learning models in this domain.

\subsubsection{GCN-based methods.}

Graph Convolutional Networks (GCNs) have emerged as an effective architecture in human action recognition due to their strong ability to model the body's bone and joint structure \cite{stgcn,shi2019twostream,chen2021channelwise,Chi_2022_CVPR}. ST-GCN~\cite{stgcn} pioneered the use of GCNs for skeleton-based action recognition, automating the capture of spatial and temporal joint dynamics without hand-crafted features. CTR-GCN~\cite{chen2021channelwise} further improved accuracy by refining the graph topology to better capture underlying joint connections, enhancing responsiveness to complex actions. Info-GCN~\cite{Chi_2022_CVPR} enriched models with deeper contextual joint information and leverages self-attention based graph convolution and multi-modal representation, leading to a more nuanced understanding of action dynamics. HD-GCN~\cite{hdgcn} built on these advancements by integrating hierarchical and dynamic graph convolutions that adapt to the varying complexity of human motions. However, GCNs face challenges in capturing long-range dependencies due to their reliance on fixed graph topologies, particularly in actions involving distant body parts or spanning numerous frames. Consequently, many recent GCN models have integrated learnable topologies and attention-like mechanisms to enhance their performance.

\subsubsection{Transformer-based methods.}

Transformer models~\cite{vaswani2017}, originating in natural language processing, have emerged as a novel architecture in skeleton-based action recognition \cite{plizzari2021skeleton,zhou2023hypergraph,long2023step}. Their self-attention mechanism offers a distinct advantage over GCNs by effectively capturing long-range dependencies and interactions between distant body parts or frames. This enables Transformers to better model complex actions involving coordinated motions across the entire body. Additionally, unlike GCNs with fixed topologies, Transformers learn adaptive representations for different body parts and actions~\cite{xin2023transformer}, enhancing their adaptability to diverse action categories. 

In~\cite{zhou2023hypergraph}, Hyperformer integrates a graph distance-based positional embedding that aligns with skeletal structures. Furthermore, it introduces a Hypergraph Self-Attention (HyperSA) mechanism to capture complex kinematic interactions, enhancing model interpretability. By strategically omitting traditional Multi-Layer Perceptron (MLP) layers, they simplify the architecture and reduce computational demands without compromising accuracy. In~\cite{long2023step}, STEP-CATFormer combines CTR-GCN~\cite{chen2021channelwise} with dual joint cross-attention Transformers, specifically designed for upper-lower and hand-foot body parts. It also designs a Super Dynamic Temporal Transformer to effectively capture complex motion patterns across temporal dimension.

Despite these strengths, current Transformers have only reached a similar level of accuracy as SOTA GCNs. Also, they incur significant computational costs during training and inference, since models with more parameters and training data tend to achieve better results~\cite{long2023step}. Ongoing research focuses on optimizing Transformer efficiency and harnessing their potential in human action recognition by tailoring them to the unique characteristics of skeleton data. 

\subsection{Model Compression and Low-Rank Approximation}

Reducing neural network size while maintaining accuracy is a well-established strategy in deep learning, offering computational advantages across various architectures~\cite{molchanov2017pruning}. Techniques like structured pruning and the identification of sparse sub-networks have been successfully employed to enhance inference speed~\cite{frankle2019lottery}. However, uniform pruning across layers has shown mixed results, sometimes requiring retraining/fine-tuning or even leading to decreased performance~\cite{lv-etal-2023-lightformer,8099498}. More recent studies suggest that selective pruning of specific weight matrices, particularly in later layers, can improve generalization without the need for additional training~\cite{sharma2023truth}.

Low-rank approximation, utilizing techniques like SVD, is another effective method for reducing model complexity, especially in Transformers~\cite{8099498}. This approach has been explored in various forms, including linear approximations of self-attention~\cite{Linformer}, progressive rank reduction with minimal fine-tuning~\cite{ProgressiveLowRank}, and hybrid sparse-low-rank approximations~\cite{Scatterbrain}. Adaptations like LoRA have gained popularity for fine-tuning large language models (LLMs) by using low-rank matrices in the updating step~\cite{hu2021lora}. While excessive rank reduction can hinder performance, strategic application of SVD, weight transfer, and parameter sharing can mitigate these effects~\cite{lv-etal-2023-lightformer}. Further research has demonstrated the effectiveness of targeting Multi-Head Self-Attention (MHSA) layers for low-rank approximation while pruning FFN layers~\cite{LoRAP2024}.

In the domain of Graph Transformers, low-rank global representation-based attention has been introduced to capture node relationships effectively~\cite{GlobalGraph}. 

Although low-rank Transformer has been studied extensively, to the best of our knowledge, its application to skeleton-based action recognition has not yet been explored in the literature.

\section{Preliminaries}

\subsection{Self-Attention}

Self-Attention (SA) mechanisms have revolutionized sequence-based modeling \cite{vaswani2017}, enabling a deeper understanding of complex dependencies. In skeleton-based action recognition, where spatial configuration and temporal dynamics are crucial, SA offers a distinct advantage by effectively capturing long-range interactions between joint positions across the entire sequence.

The core strength of SA lies in its ability to compute the representation of a position within a sequence by considering all other positions within that sequence. This mechanism can be formalized as follows \cite{vaswani2017}.
\[
\text{Attention}(\mathbf{Q}, \mathbf{K}, \mathbf{V}) = \text{softmax}\left(\frac{\mathbf{Q}\mathbf{K}^T}{\sqrt{d_k}}\right)\mathbf{V},
\]
where \( \mathbf{Q} \), \( \mathbf{K} \), and \( \mathbf{V} \) represent the Query (Q), Key (K), and Value (V) matrices, derived from the input data. The softmax function is applied to the dot products of the queries with the keys, scaled by the inverse square root of the dimensionality of the keys, \( d_k \), ensuring stable gradients during training.

Furthermore, the SA architecture enables the use of MHSA, where the input is projected multiple times through distinct, learned linear projections to produce queries, keys, and values. Each head captures different facets of the information, and their outputs are concatenated and linearly transformed, resulting in a richer representation of the sequence:
\[
\text{MultiHead}(\mathbf{Q}, \mathbf{K}, \mathbf{V}) = \text{Concat}(\text{head}_1, \dots, \text{head}_h)\mathbf{W}^O,
\]
where $\mathbf{W}^O$ is the output linear transformation matrix. Each head is computed as:
\[
\text{head}_i = \text{Attention}(\mathbf{QW}_i^Q, \mathbf{KW}_i^K, \mathbf{VW}_i^V),
\]
where $\mathbf{W}_i^Q$, $\mathbf{W}_i^K$, and $\mathbf{W}_i^V$ are the projection matrices for the queries, keys, and values, respectively, in the $i^\text{th}$ attention head.

This architecture enables the model to focus on different positions within the sequence, thus capturing varying dependencies across multiple representation subspaces at different positions.

\subsection{Low Rank Approximation}

Low-Rank Decomposition (LRD) is a key technique in model compression for reducing computational complexity, particularly in Transformer architectures. Within each MLP layer of a Transformer, the weight matrix \( \mathbf{W} \in \mathbb{R}^{C_{\text{in}} \times C_{\text{out}}} \) can be quite large, where $C_{\text{in}}$ and $C_{\text{out}}$ are the size of the input and output respectively.


The core idea of LRD is to approximate the weight matrix \( \mathbf{W} \) with a product of three smaller matrices, effectively reducing its rank and the total number of trainable parameters. This decomposition is commonly achieved using SVD. Let $n\leq\min(C_\text{in}, C_\text{out})$ be the rank of \( \mathbf{W} \). SVD decomposes \( \mathbf{W} \) into three matrices:
\[
\mathbf{W} = \mathbf{U} \mathbf{\Sigma} \mathbf{V}^T,
\]

\noindent where \( \mathbf{U} \in \mathbb{R}^{C_{\text{in}} \times C_{\text{in}}} \) and \( \mathbf{V} \in \mathbb{R}^{C_{\text{out}} \times C_{\text{out}}} \) are orthogonal matrices, and \( \mathbf{\Sigma} \in \mathbb{R}^{C_{\text{in}} \times C_{\text{out}}} \) is a diagonal matrix containing the singular values. By selecting only the largest $k \leq n$ singular values and corresponding sub-matrices in \( \mathbf{U} \) and \( \mathbf{V}^T \), the matrix \( \mathbf{W} \) can be approximated as~\cite{LRD}:
\[
\mathbf{W_{LR}} = \mathbf{U}_k \mathbf{\Sigma}_k \mathbf{V}_k^T,
\]
where \( \mathbf{W_{LR}}\in \mathbb{R}^{C_{\text{in}} \times C_{\text{out}}}\) has a lower rank \( k \). This significantly reduces the matrix rank while attempting to preserve the essential information of the original weights matrix \( \mathbf{W} \).

\section{Method}

\subsection{Low-Rank Approximation of the Fully Connected Layers}

Our approach commences by identifying fully connected layers within the Transformer blocks of two SOTA pre-trained models, Hyperformer\cite{zhou2023hypergraph} and STEP-CATFormer\cite{long2023step}. Within these layers, we target the weight matrices \( \mathbf{W} \), which significantly contribute to the computational demands of these Transformer-based architectures. We apply SVD to each identified \( \mathbf{W} \) to obtain its decomposition \( \mathbf{W} = \mathbf{U} \mathbf{\Sigma} \mathbf{V}^T \).  By selecting a reduced rank \( k \), we construct truncated matrices \( \mathbf{U}_k \), \( \mathbf{\Sigma}_k \), and \( \mathbf{V}_k^T \). Subsequently, we compute \( \mathbf{W_1} = \mathbf{U}_k \mathbf{\Sigma}_k \) and set \( \mathbf{W_2} = \mathbf{V}_k^T \), effectively decomposing the original weight matrix into two lower-rank matrices.

With the matrices \( \mathbf{W_1} \) and \( \mathbf{W_2} \) defined, the original fully connected layer \( \mathbf{W} \), with dimensions \( C_{\text{in}} \times C_{\text{out}} \), is reconfigured into two cascaded fully connected layers. The first layer, \( \mathbf{W_1} \), has dimensions \( C_{\text{in}} \times k \), and the second, \( \mathbf{W_2} \), has dimensions \( k \times C_{\text{out}} \). This reconfiguration reduces the number of parameters from \( C_{\text{in}} \times C_{\text{out}} \) to \( k \times (C_{\text{in}} + C_{\text{out}}) \), achieving a significant reduction especially when \( k \) is chosen to be small.

\subsection{Fine-tuning Strategy}

To mitigate potential information loss and performance degradation from rank reduction in the previous steps, we implement a fine-tuning strategy. The entire reconfigured low-rank model undergoes fine-tuning for a reduced number of epochs to compensate for any performance loss, which may even yield further performance gains.

For Hyperformer~\cite{zhou2023hypergraph}, we first initialize the base model with publicly available pre-trained weights and apply low-rank approximation afterwards. We then fine-tune the entire resulting model by employing a base learning rate of 0.0025 with a decay rate of 0.1 at epochs 5, 25, 45, 65, and 85. The fine-tuning process uses cross-entropy loss for 105 epochs with a batch size of 64 for both retraining and testing.

For STEP-CATFormer~\cite{long2023step}, due to lack of pre-trained weights, we train the original model from scratch for 160 epochs with a starting learning rate of 0.06, which is decreased by 0.1 at epochs 125, 135, and 150 after a 5-epoch warm-up. A larger batch size of 90 is used for training and testing due to the model's extensive parameter space. Then we apply low-rank approximation to these trained weights and initialize the low-rank model with these new weights. To fine-tune the low-rank model, a lower base learning rate of 0.001 is used, which is reduced at epochs 5, 15, 25, and 40 over 50 epochs.

\section{Experiments}

\subsection{Datasets and Experimental Settings}

\subsubsection{NTU-RGB+D.}~NTU-RGB+D~\cite{ntu} is among the most widely used for benchmarking in the field of skeleton-based action recognition. It comprises of 56,880 video samples collected from 40 distinct human subjects across 60 action classes. Each action is captured by three Microsoft Kinect V2 cameras placed at different angles, providing a diverse set of viewpoints. The dataset facilitates two primary evaluation settings: Cross-Subject (X-Sub), where models are trained and tested on distinct sets of subjects, and Cross-View (X-View), where the training and testing data are split based on camera views.

\subsubsection{NTU-RGB+D 120.}~NTU-RGB+D 120~\cite{ntu120} is an extension of the NTU-RGB+D dataset which doubles the number of action classes to 120 and includes a total of 114,480 video samples performed by 106 different participants. Similar to its predecessor, it supports two benchmark evaluation strategies: Cross-Subject and Cross-Setup. The latter divides the data based on the setup configurations, further challenging the model's generalization capabilities across different environmental contexts.

\subsubsection{Experimental Settings.}

Our experiments were conducted using the Python programming language and the PyTorch framework~\cite{paszke2019pytorch} on four NVIDIA Tesla V100 GPUs with 16GB of memory each. 

\subsection{Performance Comparison}

Table~\ref{tab1} summarizes the number of parameters, FLOPs, and performance in the two datasets for Hyperformer, STEP-CATFormer, and our low-rank approximations of the two methods. Our method significantly reduces the parameter count with similar or even better performance. These reductions translate to substantial increase in inference speed, effectively enhancing computational efficiency without compromising accuracy.

\subsubsection{Hyperformer.} The Hyperformer model consists of 10 cascaded TCN-ViT layers, each having a MHSA module~\cite{zhou2023hypergraph}. By strategically reducing the Q matrices to rank 1 and K matrices to rank 3 while preserving the original rank of V matrices (original size of \( 216\times216 \)). The model's parameter count is reduced from 2.73 M to 1.90 M, with only 1\% performance decrease, showcasing the potential for significant model compression with minimal accuracy loss. The performance can be improved after fine-tuning, as demonstrated in Table~\ref{tab1}. Notably, the Top-1 accuracy on both NTU-RGB+D datasets surpasses the original pre-trained model in the Cross-Subject setting and nearly matches in others.


\subsubsection{STEP-CATFormer.}
The STEP-CATFormer model has four large Multi Body-part Cross Attention (MBCA) blocks, each having two cascaded $6400\times6400$ fully connected layers. Additionally, an MLP layer with two cascaded fully connected layers of the same dimensions is used for processing fused temporal features~\cite{long2023step}. The STEP-CATFormer model has a significantly larger number of parameters compared to Hyperformer (417.98 million vs. 2.73 million).


After investigating different configurations, we reduce the ranks of fully connected weight matrices within the MBCA blocks to $r_{\text{MBCA}}=5$ and those in the final MLP layer to $r_{\text{MLP}}=30$. Our proposed \texttt{LORTSAR} can drastically reduce the complexity of the STEP-CATFormer model. The Model size is reduced from 417.98 M to 10.17 M (a 97.6\% reduction), and the computational load is reduced from 18.35 GFLOPs to 5.30 GFLOPs. Furthermore, after fine-tuning, the model's performance even surpasses the original accuracies in X-Sub setups for both NTU-RGB+D and NTU-RGB+D 120.

\begin{table}[t]
\caption{Performance comparison of Hyperformer and STEP-CATFormer models before applying \texttt{LORTSAR}, after applying \textbf{\texttt{LORTSAR}} before fine-tuning (denoted as \texttt{LORTSAR}\textsuperscript{(-)}), and after applying \texttt{LORTSAR} with fine-tuning by the number of parameters, FLOPs, and Top-1 accuracy on NTU-RGB+D and NTU-RGB+D 120.}\label{tab1}
\centering
\fontsize{7}{9}\selectfont
\resizebox{\textwidth}{!}{
\begin{tabular}{l||c|c|cc|cc}
\specialrule{1.5pt}{0pt}{0pt}
\hline
\multirow{2}{*}{Methods} & \multirow{2}{*}{\begin{tabular}[c]{@{}c@{}}Param.\\ (M)\end{tabular}} & \multirow{2}{*}{\begin{tabular}[c]{@{}c@{}}FLOPs\\ (G)\end{tabular}} & \multicolumn{2}{c|}{NTU-RGB+D (\%)}       & \multicolumn{2}{c}{NTU-RGB+D 120 (\%)}    \\ \cline{4-7}
                         &                                                                       &                                                                      & \multicolumn{1}{c|}{X-Sub 60} & X-View 60 & \multicolumn{1}{c|}{X-Sub 120} & X-Set 120 \\ \hline
\specialrule{1.5pt}{0pt}{0pt}
Hyperformer \cite{zhou2023hypergraph}             & 2.73                                                                  & 4.81                                                                    & \multicolumn{1}{c|}{92.69}     & \textbf{96.46}      & \multicolumn{1}{c|}{89.89}      & \textbf{91.25}      \\ \hline
Hyperformer + \texttt{LORTSAR}\textsuperscript{(-)} & 1.90 & 3.05 & \multicolumn{1}{c|}{92.22} & 95.80 & \multicolumn{1}{c|}{88.42} & 90.49 \\ \hline
Hyperformer + \texttt{LORTSAR}    & \textbf{1.90}                                                                   & \textbf{3.05}                                                                    & \multicolumn{1}{c|}{\textbf{92.84}}     & 96.44      & \multicolumn{1}{c|}{\textbf{89.93}}      & 91.21      \\ \hline
\specialrule{0.75pt}{0pt}{0pt}
STEP-CATFormer \cite{long2023step}          & 417.98                                                                   & 18.35                                                                 & \multicolumn{1}{c|}{91.77}     & \textbf{96.27}      & \multicolumn{1}{c|}{88.33}      & \textbf{89.92}      \\ \hline
STEP-CATFormer + \texttt{LORTSAR}\textsuperscript{(-)} & 10.17 & 5.30 & \multicolumn{1}{c|}{6.44} & 3.21 & \multicolumn{1}{c|}{5.38} & 19.78 \\ \hline
STEP-CATFormer + \texttt{LORTSAR} & \textbf{10.17}                                                                    & \textbf{5.30}                                                                  & \multicolumn{1}{c|}{\textbf{92.01}}     & 96.09      & \multicolumn{1}{c|}{\textbf{88.38}}      & 89.79      \\ \hline
\end{tabular}
}
\end{table}

\subsection{Ablation study}

To investigate the effects of different factors, we conduct three ablation studies, specifically targeting the type of weight matrices, the degree of rank reduction, and the fine-tuning process. We systematically vary these parameters to find the optimal trade-off between model size and performance for efficiency and accuracy.

\subsubsection{Effect of different low-rank weight matrices.}

In the Hyperformer model, according to Table~\ref{tab2}, aggressive rank reduction to 1 differentially impacts accuracy across Q, K, and V matrices within the MHSA blocks. While Q matrices demonstrate the least sensitivity, reducing V matrices to rank 1 severely impairs model functionality. However, subsequent fine-tuning can restore the accuracy, a finding that will be explored in a subsequent ablation study.

For the STEP-CATFormer model, according to Table~\ref{tab3}, our results exhibit heightened sensitivity to rank reductions in the final MLP layer compared to the MBCA blocks. This is particularly evident at lower ranks (5 and 10), where significant accuracy losses occur, highlighting the MLP layer's crucial role in integrating and classifying temporal features for accurate action recognition. However, we explore choosing the proper rank in the next ablation study section.

\noindent
\begin{table}[t]
\centering
\begin{minipage}[ht]{0.47\textwidth}
\centering
\fontsize{7}{9}\selectfont
\caption{Effect of rank selection (\textit{before fine-tuning}) on parameter count and Top-1 accuracy for Query (Q), Key (K), and Value (V) weight types in \textbf{Hyperformer} on NTU-RGB+D, X-Sub setup, joint modality.}
\label{tab2}
\vspace{0.3mm}
\resizebox{\textwidth}{!}{
\begin{tabular}{c|c|c|c}
\specialrule{1pt}{0pt}{0pt}
\hline
\textbf{Weights} & \textbf{Rank} & \textbf{Param. (M)} & \textbf{Top-1 (\%)} \\
\hline
\specialrule{1pt}{0pt}{0pt}
Original & - & 2.729 & 90.87 \\
\hline
\multirow{3}{*}{Q} & 3 & 2.321 & 90.87 \\
                   & 2 & 2.317 & 90.87 \\
                   & 1 & 2.313 & 90.84 \\
\hline
\multirow{3}{*}{K} & 3 & 2.321 & 89.94 \\
                   & 2 & 2.317 & 89.08 \\
                   & 1 & 2.313 & 86.11 \\
\hline
\multirow{3}{*}{V} & 3 & 2.321 & 7.42  \\
                   & 2 & 2.317 & 6.26  \\
                   & 1 & 2.313 & 3.87  \\
\hline
\multirow{5}{*}{Q, K} & 3, 3 & 1.913 & 89.95 \\
                        & 2, 2 & 1.905 & 89.14 \\
                          & \textbf{1, 3} & \textbf{1.905} & \hspace{2.5mm}\textbf{89.94}\hspace{1mm}\textbf{*} \\
                            & 1, 2 & 1.900 & 89.09 \\
                              & 1, 1 & 1.896 & \hspace{2.5mm}86.04\hspace{1mm}\textbf{*} \\
\hline
\multirow{3}{*}{Q, K, V} & 3, 3, 3 & 1.488 & 7.25 \\
                             & 2, 2, 2 & 1.484 & 6.19 \\
                                 & 1, 1, 1 & 1.480 & \hspace{2.5mm}3.75\hspace{1mm}\textbf{*} \\
\hline
\end{tabular}
}
\footnotetext{\scriptsize \textbf{*}\hspace{1mm}: Selected for ablation study on fine-tuning.}
\end{minipage}
\hfill
\begin{minipage}[ht]{0.47\textwidth}
\centering
\fontsize{7}{9}\selectfont
\caption{Effect of rank selection (\textit{before fine-tuning}) on parameter count and Top-1 accuracy for $r_{\text{MBCA}}$ and $r_{\text{MLP}}$ in \textbf{STEP-CATFormer} on NTU-RGB+D, X-Sub setup, joint modality.}
\label{tab3}
\resizebox{\textwidth}{!}{
\begin{tabular}{cc|c|c}
\specialrule{1pt}{0pt}{0pt}
\hline
\multicolumn{2}{c|}{\textbf{Rank}}                & \multicolumn{1}{l|}{\multirow{2}{*}{\textbf{Param. (M)}}} & \multicolumn{1}{l}{\multirow{2}{*}{\textbf{Top-1 (\%)}}} \\ \cline{1-2}
\multicolumn{1}{c|}{$r_{\text{MBCA}}$}             & $r_{\text{MLP}}$ & \multicolumn{1}{l|}{}                                     & \multicolumn{1}{l}{}                                     \\ \hline
\specialrule{1pt}{0pt}{0pt}
\multicolumn{2}{c|}{Original}                     & 417.98                                                    & 90.58                                                    \\ \hline
\multicolumn{1}{c|}{\multirow{4}{*}{5}}  & 5      & 9.02                                                      & 1.67                                                     \\
\multicolumn{1}{c|}{}                    & 10     & 9.15                                                      & 1.67                                                     \\
\multicolumn{1}{c|}{}                    & 20     & 9.40                                                      & 1.67                                                     \\
\multicolumn{1}{c|}{}                    & 30     & 9.66                                                      & \hspace{2.5mm}1.67\hspace{1mm}\textbf{*}                                                     \\ \hline
\multicolumn{1}{c|}{\multirow{4}{*}{10}} & 5      & 9.53                                                      & 1.67                                                     \\
\multicolumn{1}{c|}{}                    & 10     & 9.66                                                      & 1.67                                                     \\
\multicolumn{1}{c|}{}                    & 20     & 9.91                                                      & 2.64                                                     \\
\multicolumn{1}{c|}{}                    & 30     & 10.17                                                     & \hspace{2.5mm}2.66\hspace{1mm}\textbf{*}                                                     \\ \hline
\multicolumn{1}{c|}{\multirow{4}{*}{20}} & 5      & 10.55                                                     & 1.67                                                     \\
\multicolumn{1}{c|}{}                    & 10     & 10.68                                                     & \hspace{2.5mm}1.67\hspace{1mm}\textbf{*}                                                     \\
\multicolumn{1}{c|}{}                    & 20     & 10.94                                                     & 9.41                                                     \\
\multicolumn{1}{c|}{}                    & 30     & 11.19                                                     & \hspace{2.5mm}34.18\hspace{1mm}\textbf{*}                                                    \\ \hline
\multicolumn{1}{c|}{\multirow{4}{*}{30}} & 5      & 11.58                                                     & 1.64                                                     \\
\multicolumn{1}{c|}{}                    & 10     & 11.71                                                     & \hspace{2.5mm}1.67\hspace{1mm}\textbf{*}                                                     \\
\multicolumn{1}{c|}{}                    & 20     & 11.96                                                     & \hspace{2.5mm}18.37\hspace{1mm}\textbf{*}                                                    \\
\multicolumn{1}{c|}{}                    & 30     & 12.22                                                     & \hspace{2.5mm}48.81\hspace{1mm}\textbf{*}                                                    \\ \hline
\end{tabular}
}
\footnotetext{\scriptsize \textbf{*}\hspace{1mm}: Selected for ablation study on fine-tuning.}
\end{minipage}
\end{table}

\subsubsection{Effect of different ranks.}

We study the effect of varying ranks across Q, K, and V weight matrices in Hyperformer, and MBCA blocks and the final MLP layer in STEP-CATFormer. For Hyperformer, we vary the ranks of (1, 2, 3) accross Q, K, and V weight matrices. Our results reveal that higher ranks retain higher accuracy as outlined in Table~\ref{tab2}, with reductions in V matrices proving most detrimental. The optimal configuration identified for further fine-tuning has rank 1 for Q and rank 3 for K matrices (Table~\ref{tab4}).

For the STEP-CATFormer model, we investigate the impact of various rank settings for the MBCA blocks (\(r_\text{MBCA}\)) and the MLP layer (\(r_\text{MLP}\)), as summarized in Table~\ref{tab3}. Lower ranks for \(r_\text{MBCA}\) (5 or 10) significantly impair performance, regardless of the \(r_\text{MLP}\) value. Higher ranks for \(r_\text{MBCA}\) (20 or 30) generally perform better, but lower \(r_\text{MLP}\) values still hinder performance. Notably, configurations with both \(r_\text{MBCA}\) and \(r_\text{MLP}\) above 20 exhibit minimal performance loss. Based on these findings, several configurations, particularly those with higher ranks, are selected for further fine-tuning in subsequent ablation studies section, as shown in Figure~\ref{fig1}.

\noindent
\begin{table}[t]
\centering
\begin{minipage}[ht]{0.35\textwidth}
\centering
\fontsize{7}{10}\selectfont
\caption{Highest Top-1 Accuracy and corresponding training epoch after fine-tuning \textbf{Hyperformer}.}
\label{tab4}
\vspace{1mm}
\begin{tabular}{l|l|l}
\specialrule{1pt}{0pt}{0pt}
\hline
\textbf{Layers}         & \textbf{Epoch} & \textbf{Top-1 (\%)} \\
\specialrule{1pt}{0pt}{0pt}
\hline
Original & -          & 90.87           \\ \hline
\textbf{Q1}\textsuperscript{*}, \textbf{K3}         & 39         & \textbf{90.94}           \\ \hline
Q1, K1         & 104        & 90.54           \\ \hline
Q1, K1, V1     & 96         & 87.01           \\ \hline
Q1, K1, V3     & 56         & 89.07           \\ \hline
\end{tabular}
\footnotetext{\scriptsize * Q1\hspace{0.3mm}: \texttt{LORTSAR} with rank 1 on Query matrices.}
\end{minipage}
\hfill
\begin{minipage}[ht]{0.6\textwidth}
\centering
\fontsize{8}{10}\selectfont
\begin{tikzpicture}[scale=1.05]
    \begin{axis}[
        width=\textwidth,
        height=0.8\textwidth,
        xlabel={Epoch},
        xlabel style={yshift=0.5ex},
        ylabel={Top-1 acc. (\%)},
        ylabel style={yshift=-1ex},
        xmin=0, xmax=30,
        ymin=0, ymax=93,
        grid=major,
        legend style={
            font=\tiny,
            at={(0.98,0.43)},
            anchor=east
        },
        legend cell align={left}
    ]

    \addplot[color=Green, thick, dashdotted] coordinates {
    (0,48.81) (1,89.69) (2,89.89) (3,89.88) (4,90.09) (5,90.22) 
    (6,90.19) (7,90.02) (8,90.11) (9,90.31) (10,90.20)
    (11,90.15) (12,90.17) (13,90.16) (14,90.22) (15,90.27)
    (16,90.20) (17,90.22) (18,90.22) (19,90.27) (20,90.23)
    (21,90.25) (22,90.28) (23,90.30) (24,90.22) (25,90.23)
    (26,90.18) (27,90.11) (28,90.23) (29,90.17) (30,90.25)
    (31,90.20) (32,90.14) (33,90.23) (34,90.19) (35,90.34)
    (36,90.14) (37,90.28) (38,90.32) (39,90.25) (40,90.14)
    (41,90.22) (42,90.16) (43,90.23) (44,90.23) (45,90.25)
    (46,90.21) (47,90.26) (48,90.30) (49,90.18) (50,90.30)
    };
    \addlegendentry{30-30}

    \addplot [thick, color=Dandelion, dashed] coordinates {
    (0,1.67) (1,19.57) (2,44.70) (3,72.79) (4,85.22) (5,87.41) (6,88.08) (7,88.46) (8,88.57) (9,88.51) (10,88.63) (11,88.61) (12,88.83) (13,88.88) (14,88.86) (15,88.65) (16,88.78) (17,88.71) (18,88.92) (19,88.96) (20,88.95) (21,88.80) (22,88.66) (23,88.96) (24,88.77) (25,88.96) (26,88.85) (27,88.83) (28,88.81) (29,88.91) (30,88.94) (31,88.80) (32,88.91) (33,88.72) (34,88.88) (35,88.85) (36,88.84) (37,88.77) (38,88.82) (39,88.80) (40,88.86) (41,88.74) (42,88.72) (43,88.95) (44,88.92) (45,88.86) (46,88.91) (47,88.80) (48,88.74) (49,88.85) (50,88.69)
};
\addlegendentry{30-10}
    
    \addplot[color=Purple, thick, dashed] coordinates {
    (0,34.18) (1,89.69) (2,89.90) (3,90.13) (4,90.27) (5,90.08) 
    (6,90.32) (7,90.24) (8,90.24) (9,90.40) (10,90.20)
    (11,90.26) (12,90.31) (13,90.30) (14,90.23) (15,90.23)
    (16,90.34) (17,90.18) (18,90.36) (19,90.21) (20,90.23)
    (21,90.27) (22,90.31) (23,90.31) (24,90.33) (25,90.30)
    (26,90.12) (27,90.25) (28,90.28) (29,90.30) (30,90.29)
    (31,90.25) (32,90.18) (33,90.14) (34,90.27) (35,90.19)
    (36,90.22) (37,90.22) (38,90.27) (39,90.27) (40,90.11)
    (41,90.20) (42,90.16) (43,90.28) (44,90.12) (45,90.30)
    (46,89.88) (47,90.30) (48,90.20) (49,90.17) (50,90.19)
};
\addlegendentry{20-30}

    \addplot[color=Gray] coordinates {
    (0,1.67) (1,17.75) (2,41.78) (3,70.07) (4,84.95) (5,87.38)
    (6,88.05) (7,88.58) (8,88.51) (9,88.66) (10,88.58)
    (11,88.63) (12,88.62) (13,88.80) (14,88.59) (15,88.71)
    (16,88.78) (17,88.63) (18,88.75) (19,88.51) (20,88.74)
    (21,88.74) (22,88.71) (23,88.82) (24,88.86) (25,88.67)
    (26,88.79) (27,88.79) (28,88.74) (29,88.70) (30,88.63)
    (31,88.66) (32,88.81) (33,88.80) (34,88.70) (35,88.66)
    (36,88.72) (37,88.81) (38,88.72) (39,88.80) (40,88.82)
    (41,88.81) (42,88.82) (43,88.62) (44,88.79) (45,88.62)
    (46,88.63) (47,88.69) (48,88.75) (49,88.80) (50,88.83)
};
\addlegendentry{20-10}

    \addplot[color=Brown] coordinates {
    (0, 2.66) (1, 89.48) (2, 89.65) (3, 90.09) (4, 89.86) (5, 90.00) (6, 90.14) (7, 90.17) (8, 90.14) (9, 90.19) (10, 90.19) (11, 90.24) (12, 90.26) (13, 90.14) (14, 90.10) (15, 90.08) (16, 90.28) (17, 90.17) (18, 90.28) (19, 89.77) (20, 90.35) (21, 90.15) (22, 90.21) (23, 90.33) (24, 90.25) (25, 90.19) (26, 90.21) (27, 90.32) (28, 90.36) (29, 90.23) (30, 90.24) (31, 90.29) (32, 90.25) (33, 90.33) (34, 90.29) (35, 90.13) (36, 90.14) (37, 90.30) (38, 90.27) (39, 90.23) (40, 90.28) (41, 90.25) (42, 90.30) (43, 90.35) (44, 90.36) (45, 90.30) (46, 90.26) (47, 90.30) (48, 90.26) (49, 90.30) (50, 90.26)
    };
    \addlegendentry{10-30}

    \addplot[color=red, thick, dashdotted, mark=square, mark repeat=5, mark options={solid}, mark size=1.3pt] coordinates {
    (0,1.67) (1,89.19) (2,89.68) (3,89.80) (4,89.96) (5,90.03) 
    (6,90.12) (7,89.77) (8,90.06) (9,90.13) (10,90.02)
    (11,90.07) (12,90.07) (13,90.17) (14,90.10) (15,90.10)
    (16,90.15) (17,90.10) (18,90.19) (19,90.16) (20,90.19)
    (21,90.19) (22,90.16) (23,90.18) (24,90.08) (25,90.29)
    (26,90.12) (27,90.24) (28,90.38) (29,90.17) (30,90.18)
    (31,90.06) (32,90.11) (33,90.19) (34,90.16) (35,90.11)
    (36,90.08) (37,90.09) (38,90.17) (39,90.05) (40,90.13)
    (41,90.15) (42,90.10) (43,90.10) (44,90.12) (45,90.16)
    (46,90.11) (47,90.00) (48,90.11) (49,90.16) (50,90.22)
    };
    \addlegendentry{\textbf{5-30 *}}

    \addplot[color=Cyan, thick] coordinates {
    (0,1.67)(1,2.14) (2,2.20) (3,3.57) (4,10.31) (5,38.50) 
    (6,72.37) (7,82.20) (8,82.94) (9,83.07) (10,83.36)
    (11,83.56) (12,83.82) (13,84.04) (14,84.04) (15,84.02)
    (16,84.35) (17,84.25) (18,84.35) (19,84.30) (20,84.20)
    (21,84.10) (22,84.19) (23,84.21) (24,84.39) (25,84.29)
    (26,84.35) (27,84.63) (28,84.29) (29,84.30) (30,84.53)
    (31,84.36) (32,84.32) (33,84.52) (34,84.31) (35,84.33)
    (36,84.48) (37,84.40) (38,84.28) (39,84.40) (40,84.45)
    (41,84.29) (42,84.45) (43,84.50) (44,84.16) (45,84.37)
    (46,84.44) (47,84.31) (48,84.35) (49,84.37) (50,84.32)
    };
    \addlegendentry{5-5}

    \addplot[color=Thistle, thick] coordinates {
        (0,1.67) (1,1.67) (2,1.67) (3,1.67) (4,1.67) (5,1.67) (6,1.67) (7,1.67) (8,1.67) (9,1.67) (10,1.67) 
        (11,1.67) (12,1.67) (13,1.67) (14,1.67) (15,1.67) (16,1.67) (17,1.67) (18,1.67) (19,1.67) (20,1.67) 
        (21,1.67) (22,1.67) (23,1.67) (24,1.67) (25,1.67) (26,1.67) (27,1.67) (28,1.67) (29,1.67) (30,1.67) 
        (31,1.67) (32,1.67) (33,1.67) (34,1.67) (35,1.67) (36,1.67) (37,1.67) (38,1.67) (39,1.67) (40,1.67) 
        (41,1.67) (42,1.67) (43,1.67) (44,1.67) (45,1.67) (46,1.67) (47,1.67) (48,1.67) (49,1.67) (50,1.67)
    };
    \addlegendentry{1-1}

    \addplot[dashed, thick] coordinates {(0,90.58) (50,90.58)};
    \addlegendentry{original}

    \end{axis}
\end{tikzpicture}
\vspace{2mm}

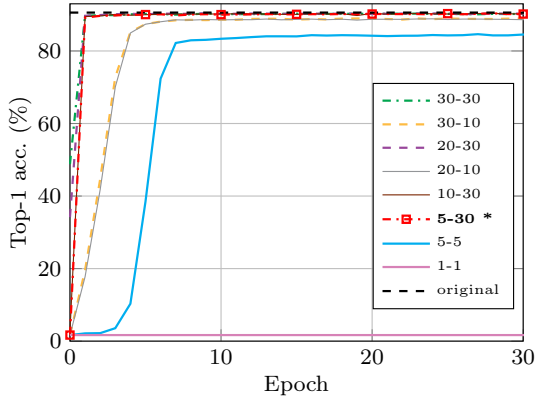
\captionof{figure}{Effect of fine-tuning on Top-1 accuracy of \textbf{STEP-CATFormer} with \texttt{LORTSAR} at different ranks. Each \texttt{LORTSAR} rank configuration is denoted as ($r_{\text{MBCA}}-r_{\text{MLP}}$)}
\label{fig1}
\end{minipage}
\vspace{-1mm}
\end{table}

\subsubsection{Effect of fine-tuning.}

This ablation study examines the impact of fine-tuning the Hyperformer model after applying the highlighted low-rank configurations in Table~\ref{tab4}. The modified hyperparameters are used because the original model was pre-trained to a high accuracy. Our results reveal that fine-tuning consistently improves performance, with some fine-tuned models surpassing the original accuracy. Although performance converges after several epochs for certain modalities, no consistent pattern emerges across modalities to justify early stopping. In addition, accuracy generally decreases with parameter count across configurations. However, this minor decrease may be acceptable given the significant parameter reduction. Notably, configurations with reduced V matrices achieve satisfactory performance after fine-tuning, despite their initially low accuracy. This indicates that all MHSA matrices (Q, K, V) are suitable for SVD-based compression, warranting further exploration.

For STEP-CATFormer, we apply various rank configurations detailed in Table~\ref{tab3} and Figure~\ref{fig1}. While the highest compression setting (\(r_\text{MBCA} = 1\) and \(r_\text{MLP} = 1\)) shows no improvement from fine-tuning, all other configurations exhibit notable enhancements. The extent of accuracy recovery via fine-tuning correlates with the level of rank reduction. As depicted in Figure~\ref{fig1}, except for the configurations \((5,5)\), \((20,10)\), and \((30,10)\), fine-tuning successfully restores performance losses. Consequently, we identify the configuration \((r_\text{MBCA}, r_\text{MLP}) = (5,30)\) as optimal, balancing maximum model compression with full recovery of performance through fine-tuning.

\section{Discussions and Conclusions}

In this paper, we investigate applying low-rank Transformer to two SOTA Transformer models in skeleton-based action recognition: Hyperformer and STEP-CATFormer. Our results demonstrate that significant reduction in model parameters is achievable with minimal accuracy loss or even performance gains through careful selection of weight matrix types and the corresponding ranks. Fine-tuning consistently improves accuracy, even in models with substantial compression. Our findings highlight two key benefits of \texttt{LORTSAR} in compressing Transformer models: \textbf{1)} Most SOTA Transformer models can be compressed, offering a favorable accuracy-parameter trade-off and enabling deployment in resource-limited settings. Further research into alternative compression methods or fine-tuning strategies may yield additional efficiency and accuracy gains. \textbf{2)} Fine-tuning consistently improves accuracy, suggesting its potential as a preliminary or intermediate step in training large Transformer models. By strategically balancing performance gains with computational constraints, our approach enables the development of resource-efficient Transformer architectures for skeleton-based human action recognition.


%
%
%
\bibliographystyle{splncs04}
\bibliography{LORTSAR}
\end{document}